# Enumerating Markov Equivalence Classes of Acyclic Digraph Models

**Steven B. Gillispie**
Department of Radiology
University of Washington, Box 356004
Seattle, WA 98195-6004

**Michael D. Perlman**
Department of Statistics
University of Washington, Box 354322
Seattle, WA 98195-4322

## Abstract

Graphical Markov models determined by acyclic digraphs (ADGs), also called directed acyclic graphs (DAGs), are widely studied in statistics, computer science (as Bayesian networks), operations research (as influence diagrams), and many related fields. Because different ADGs may determine the same Markov equivalence class, it long has been of interest to determine the efficiency gained in model specification and search by working directly with Markov equivalence classes of ADGs rather than with ADGs themselves. A computer program was written to enumerate the equivalence classes of ADG models as specified by Pearl & Verma's equivalence criterion. The program counted equivalence classes for models up to and including 10 vertices. The ratio of numbers of classes to ADGs appears to approach an asymptote of about 0.267. Classes were analyzed according to number of edges and class size. By edges, the distribution of number of classes approaches a Gaussian shape. By class size, classes of size 1 are most common, with the proportions for larger sizes initially decreasing but then following a more irregular pattern. The maximum number of classes generated by any undirected graph was found to increase approximately factorially. The program also includes a new variation of orderly algorithm for generating undirected graphs.

## 1   INTRODUCTION

One of the central ideas of statistical science is the assessment of dependencies among stochastic variables. *Graphical Markov models* (GMMs) use *graphs*, either undirected, directed, or mixed, to represent multivariate dependencies in a parsimonious and computationally efficient manner. A GMM is constructed by specifying *local* dependencies for each variable, or vertex, of the graph in terms of its immediate neighbors, parents, or both, yet can represent a highly varied and complex system of multivariate dependencies by means of the *global* structure of the graph. The local specification permits efficiencies in modeling, inference, and probabilistic calculations.

*Acyclic digraphs* (ADGs), also called directed acyclic graphs (DAGs), provide a particularly elegant framework for statistical modelling and analysis, both Bayesian and non-Bayesian (e.g., Lauritzen & Spiegelhalter 1988, Lauritzen *et al* 1990, Spiegelhalter & Lauritzen 1990). The recursive factorization form of the likelihood function associated with a given ADG model allows explicit Bayesian and likelihood inference for Gaussian and multinomial models, and also facilitates ADG model specification via elicitation of expert opinion. For these reasons, ADG models have been widely studied in statistics, computer science (as Bayesian networks), operations research (as influence diagrams), and many related fields (cf. Pearl 1988, Shachter & Kenley 1989, Neapolitan 1990, Whittaker 1990, Spiegelhalter *et al* 1993, Edwards 1995, Jensen 1996, Lauritzen 1996).

If no single ADG model is specified, model search procedures are necessary. Bayesian model selection algorithms seek out the ADG models with highest posterior probability, and subsequent inference proceeds conditionally on these selected models (Cooper & Herskovits 1992, Buntine 1994, Heckerman *et al* 1994, Madigan & Raftery 1994). Non-Bayesian model selection methods are similar, replacing posterior model probabilities by, for example, penalized maximum likelihoods (Chickering 1995).

Heckerman *et al* (1994) highlighted a fundamental problem with this general approach. Several different ADGs may determine the same statistical model, i.e., may determine the same set of conditional independence restrictions among a given set of random variates, hence cannot be distinguished on the basis of data alone regardless of sample size. Thus, the collection of all possible ADGs for a given set of variates naturally coalesces into one or more classes of Markov-equivalent ADGs, where all ADGs within a Markov equivalence class determine the same statistical model.

In the worst case, the Markov equivalence class containing a given ADG may be superexponentially large. For example, if the *skeleton* (the underlying undirected graph) of the ADG is complete, then the equivalence class contains exactly $n!$ ADGs. Model search and selection algorithms that ignore Markov equivalence may therefore be extremely inefficient. Treating each Markov equivalence class as a single model would overcome these difficulties.

An *essential graph* (Andersson, Madigan, & Perlman 1997) is a graphical representation of a Markov equivalence class that can have both directed and undirected edges. Each directed edge in the essential graph is oriented in the same direction in all members of the equivalence class. Each of the undirected edges will appear in different directions in at least two of the members of the class. Thus, the number and configuration of the undirected edges in an essential graph dictates the size of the associated Markov equivalence class.

In a graph, let $a$, $b$, and $c$ be different vertices. A *v-configuration* is defined when $a$ is connected to $b$ and $b$ to $c$, but $a$ is not connected to $c$. Given the same v-configuration, an *immorality* is defined when the two edges are directed so that both point toward the central $b$ vertex. Verma & Pearl (1990, 1992) showed that two ADGs are Markov-equivalent if and only if they have the same skeleton and same immoral configurations. Using this characterization, several researchers have developed procedures for searching directly over the space of Markov equivalence classes (cf. Meek 1995, Chickering 1996, Madigan *et al* 1996).

It is of substantial interest, therefore, to determine the efficiency that can be gained in model specification and search by working directly with Markov equivalence classes of ADGs rather than with ADGs themselves. A fundamental question is simply enumerative: for $n$ variates, what is the ratio $r_n$ of the number of Markov equivalence classes to the number of ADGs? If this ratio is small, especially if $r_n$ approaches 0 as $n$ becomes very large, then substantial computational savings might be achieved.

At this time, no formula is known for either the number of Markov equivalence classes or for their ratio $r_n$ to the number of ADGs. Robinson (1971) found a recursive formula for the number of ADGs on $n$ vertices; thus, given either of the two unknown quantities above the other can be computed. The enumerative question has been partially addressed by two independent researchers (Andersson, Madigan, & Perlman 1997; Volf & Studeny´ 1999). Both found the number of equivalence classes for $n \le 5$ vertices via a combination of manual and computer methods. Neither study was able to make any predictions about the values for $n \ge 6$. Steinsky (2000) found a recursive formula for the number of equivalence classes of size 1 only, and using a computer was able to calculate the ratio of the number of size 1 classes to total ADGs for $n$ up to 200. The ratio appears to be approaching a limit, thus suggesting an asymptote of approximately 0.07325, and therefore a lower bound for $r_n$. We report here the results of another attempt via computer to address this problem that computed the two quantities above for all $n \le 10$.

In what follows, we use the standard definitions for *graph* (always undirected) and *digraph* (directed graph), where no more than one edge is allowed between any two vertices and no edge is allowed to begin and end on the same vertex (no loops). The number of vertices (variates) will be symbolized as $n$. An *acyclic digraph* is a digraph where no directed path exists that leads from any vertex back to that same vertex. To *orient* a graph is to convert its undirected edges to directed ones. A *labeled* graph or digraph is one where each vertex is uniquely identified. An *unlabeled* graph or digraph is one where the identifiers have been removed. In general, there are more labeled graphs or digraphs than unlabeled ones, since removing the labels can make previously different graphs become indistinguishable. The *degree* of a vertex is the count of the number of edges connected to it. The terms *skeleton*, *v-configuration*, and *immorality* are as defined above. A *complete bipartite graph* is a graph whose set of vertices have been divided into two distinct subsets and where none of the vertices within each subset are connected to each other but every vertex in each subset is connected to every vertex in the other subset. A *complete multipartite graph* is defined similarly, but with the set of vertices possibly divided into more than two distinct subsets.

## 2    COMPUTER PROGRAM

### 2.1  ALGORITHMS

The goal of this investigation was to not only count the number of equivalence classes, but also to record their sizes. Thus the program needed to generate all ADGs and then sort them into their respective equivalence classes. According to the Verma & Pearl theorem (1990, 1992), no two ADGs in the same class can have different skeletons, therefore the work of the program could be performed on each skeleton separately. The first step then was to generate all of the labeled undirected graphs. But as these grow large in number very quickly, the program instead used unlabeled graphs. By multiplying the numeric results obtained for each unlabeled graph by the number of non-isomorphic ways the graph could be labeled, the program's effective speed was greatly increased.

Methods for computing the number of unlabeled graphs have been found by Redfield (1927) and Pólya (1937). For graphs with 10 or fewer vertices, these numbers have been computed algebraically (Oberschelp 1967) and by computer (Stein & Stein 1967). The numerical results served as checks on the correctness of the program, but these methods only count the graphs and do not generate them.

The unlabeled graphs were generated using an "orderly algorithm" as first described by Read (1978). An orderly algorithm generates a list of non-equivalent items in problems where isomorphic items must be eliminated, but does so without requiring comparison to the list of already generated items. This can greatly increase the speed of generating the list. The algorithm requires: (1) a canonical configuration for each item to be counted; (2) an ordering on the canonical configurations; and (3) an augmenting operation to create new items from a previous list in a generating sequence. In addition, three conditions on the augmenting operation, the canonical configurations, and the list order must be satisfied. The method used here generates a sequence of lists of graphs by number of edges (edge-augmentation), starting with zero edges (the single graph of $n$ vertices). The coding of the graphs uses the upper-triangular portion of the adjacency matrix, appending the 0/1 bits listed by columns from 1 to $n$ to form a binary integer. The canonical configuration is that which generates the largest integer, the list ordering is from largest to smallest, and the augmentation operation is to replace subsequent 1s into the last sequence of 0s in the coding (if one exists), starting with the largest 0. These methods together meet the necessary conditions for an orderly algorithm, according to the general problem theorem proved in Read (1978). Parts of this scheme are not new (Read 1978; Colbourn & Read 1979), but this combination and the addition of the following rules greatly increase the speed of the canonical configuration testing and do not appear to have been described before. They follow easily by considering the adjacency matrix encoded as a binary number as just described and noting that otherwise a permutation of vertex labels would place more bits in higher columns of the adjacency matrix, or higher bits within the same column, generating a larger coding.

1) The degree of the $n$th vertex must be maximal among the vertices.

2) All zero-degree vertices must be in the lowest-numbered columns.

3) The $(n-1)$th vertex must have maximal degree among all of the vertices connected to the $n$th vertex; in cases of equality, it must have maximal degree when counting all other connected vertices.

4) In a breadth-first listing of the vertex labels starting from the $n$th vertex, when a vertex is first encountered there should be no gap from highest to lowest in the current list of vertices.

Each of these can be verified simply by scanning the cells of the adjacency matrix. In case of condition 4, this amounts to finding no 0 gaps in the vertical list of bits starting from the highest vertex and proceeding first vertically upward in vertex number and then successively horizontally to the left and vertically upward in vertex number. When none of these rules immediately eliminate a non-canonical configuration, the configuration must be tested explicitly by permutation of its vertices. But again, by hierarchically considering the four vertices from highest to lowest in light of the above four rules, the number of permutations required for testing can be greatly reduced.

To calculate the number of non-isomorphic labellings for each graph, the size of the automorphism group (as represented by labellings) for the graph was first determined. Representing relabellings of the graph as members of an algebraic permutation group, the automorphism group consists of the set of permutations which reproduce the original graph. This set of permutations was found by generating all possible permutations of labellings within all subsets of vertices having the same degree and then simply checking for duplicates according to the adjacency coding. Since this number generally involved only products of small factorials, the process proceeded quickly enough. Once the size of the automorphism group was known, the number of non-isomorphic labellings could then be computed using Lagrange's Theorem by dividing the automorphism group size into $n!$. This is because the automorphism group is a subgroup of the group of all permutations; since this larger group has size $n!$ and since each coset of the automorphism group (a set of permutations producing the same result) is represented by a different graph labelling, the result is obtained. (See, for example, Hall (1976)).

Since the distribution of graphs by edges is symmetric and the number of non-isomorphic labellings is identical for a graph and its complement, it was only necessary to generate the lower half of the set of unlabeled graphs by these methods and then produce the upper half by complementation. To facilitate the subsequent analysis, the generated graphs were saved in separate computer files according to both number of vertices and number of edges. In all, these three steps (graph generation, counting labellings, and complementation) together represented less than 2% of the total computer time required for the computations for the larger vertex sets.

By the time the number of vertices has reached 10, the percentage of directed graphs that are cyclic has climbed to 99.9%. Thus, generating all directed graphs and discarding the cyclic ones is impractical. An algorithm to directly generate all ADGs for a given graph has been published by Barbosa & Szwarcfiter (1999) and was used in the program. As each graph was initially read in for processing, a scan was performed to locate its v-configurations. These represented possible positions for immoralities. Each possible position was encoded as a binary bit, which enabled the unique coding of every equivalence class on the skeleton due to the Verma & Pearl (1990, 1992) theorem. The maximum number of bits required was $n(n-1)(n-2)/6$, representing the maximum possible number of choices of three different vertices, though the actual number encountered was generally smaller. Then, as each ADG was generated, these positions only were checked for the existence of an immorality according to the current orientation of the edges and the equivalence class code was computed.

This unique code was then used as a key into a red-black binary tree (e.g. Cormen, Leiserson, & Rivest 1990). [A red-black tree is a binary search tree that is kept balanced (thus ensuring $O(\log N)$ Search and Insert operations) by use of an additional color bit (red or black) at each node and certain rules maintained by operations in the Insert procedure.] A new node was created whenever a new code appeared and its class size count was incremented whenever a repeat appearance of the same code occurred. This made possible the acquisition of total number of ADGs, total number of classes and their distributions according to class size, as well as the maximum number of v-configurations and classes produced per skeleton. These were all recorded when the program ended. By changing the input set of skeletons given to the program, analyses could be done separately according to both number of vertices and number of edges. The total computer time required, using a mid-1990s-era, midrange minicomputer was nearly 2253 CPU hours for $n=10$, almost 7 hours for $n=9$, and less than 3 minutes for $n=8$; smaller vertices required only seconds or less of CPU time.

## 2.2 VALIDATION

The program was implemented in stages, beginning with a very simplistic but slow algorithm that replicated the known results previously obtained by other researchers (Andersson, Madigan, & Perlman 1997; Volf & Studeny' 1999). Parts of the program were then progressively replaced with more sophisticated algorithms, but each time only replacing just enough to make the program run fast enough to achieve the next greater number of vertices, while still matching the previously found results. This 'bootstrapping' method helped provide assurance that the program was always working correctly, and also means that many of the smaller numbers were computed using multiple different algorithms. (The description above is of the final algorithm.) At each step of the program, obtained results were compared with previously known results. These included the number of unlabeled graphs (in total and by number of edges), the total number of labeled graphs by edges (multiplying by the number of computed labellings), the number of ADGs (in total and by number of edges), and the total number of equivalence classes of size 1.

## 3   RESULTS

The first result is simply the total number of equivalence classes by number of vertices. Table 1 shows these totals, along with the ratios of the numbers of classes to ADGs and the ratios of the counts of size 1 classes to the total class counts.

Table 1: Equivalence Class Counts

| $n$ | Equivalence classes | Cl/ADG | $Cl_1$/Cl |
|---|---|---|---|
| 1 | 1 | 1.00000 | 1.00000 |
| 2 | 2 | 0.66667 | 0.50000 |
| 3 | 11 | 0.44000 | 0.36364 |
| 4 | 185 | 0.34070 | 0.31892 |
| 5 | 8782 | 0.29992 | 0.29788 |
| 6 | 1067825 | 0.28238 | 0.28667 |
| 7 | 312510571 | 0.27443 | 0.28068 |
| 8 | 212133402500 | 0.27068 | 0.27754 |
| 9 | 326266056291213 | 0.26888 | 0.27590 |
| 10 | 1118902054495975141 | 0.26799 | 0.27507 |

Figure 1 shows the classes/ADGs ratios plotted against the number of vertices. A crude fit of this curve shows that it can be closely approximated by an exponential curve with an asymptotic value of around 0.27. This can be compared with Steinsky's suggested asymptotic ratio of the size 1 equivalence classes to total ADGs of about 0.07325: if the size 1 classes comprise 7% of the total ADGs then all of the classes together being 27% of the ADG totals is not unreasonable.

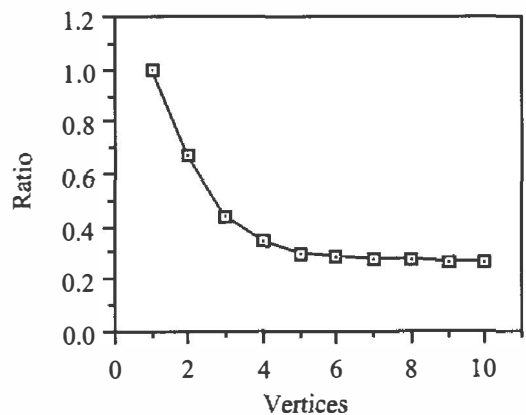

Figure 1: Ratios of Classes to ADGs by Vertices

A closer examination of the curve in Figure 1 shows that it is not exactly exponential. If it were, the curve could be modeled by the equation $a/s^n + r$, with $s$ a constant. Letting $r_n$ be the ratio on $n$ vertices of equivalence classes to ADGs, this produces the recursive formula $r_{n+1} = r_n - (r_{n-1} - r_n)/s$. However, the numbers in Table 1 show that $s$ is not constant and itself is approximately exponentially approaching an asymptotic value of 2. Even though the following coefficients have been selected partly for their simplicity, $s$ can nevertheless be predicted reasonably well by the equation $s_n = 2 + 20/3 \exp(-n/2)$. Using this formula for $s_n$ and the recursive formula above for $r_{n+1}$, initialized with $r_9$ and $r_{10}$, the value of $r_n$ can be computed for any desired $n$ and suggests that $r_n$ may be approaching an asymptotic value of about 0.26714. But due to the uncertainty in selecting the model constants, only the first three significant digits are probably accurate, producing a conservative estimate of 0.267. In any event, while no theoretical proof yet exists that any of these numbers do actually approach asymptotic limits, these results suggest that the ratio of numbers of equivalence classes to ADGs does not approach zero as $n$ approaches infinity but, instead, the two quantities scale identically as $n$ becomes large.

The numbers also suggest an asymptotic value of around 0.274 for the ratio of counts of size 1 classes to total equivalence classes: a sizeable fraction. In other words, a significant percentage of equivalence classes have only a single member.

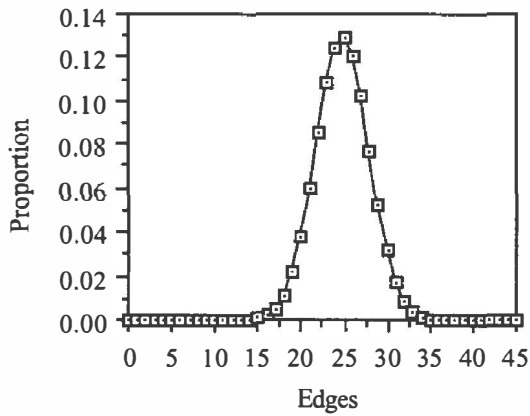

Figure 2: Equivalence Classes by Edges $(n=10)$

As it was possible for the program to count the equivalence classes separately by number of edges, Figure 2 shows the distribution of classes by edges for $n=10$. The shape matches a discretized Gaussian with a chi-squared difference of 0.000162. As the distribution appears Gaussian-like, its mean should approach its median. Except for $n=4$, the median is predicted by the formula floor$(n/2)$*ceil$(n/2)$ [where floor$(x)$ is the largest integer less than or equal to $x$ and ceil$(x)$ is the smallest integer greater than or equal to $x$ (its ceiling); they can also be thought of as rounding $x$ down or up] or, equivalently, the maximum value of $i(n-i)$ for $i$ an integer. This suggests that not only can one predict the total number of equivalence classes for large $n$, but that it may also be possible to predict them by number of edges as well.

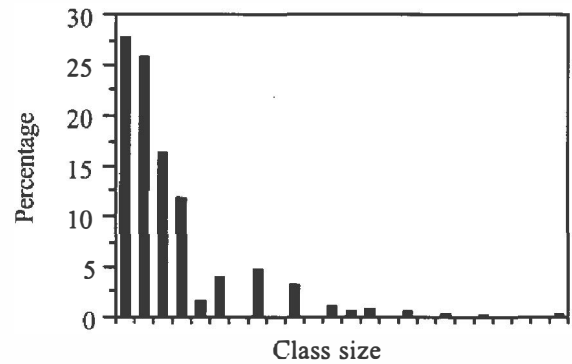

Figure 3: Classes by Size (sizes 1-24; $n=10$)

The size of each equivalence class was also recorded by the program. Figure 3 shows a histogram of the percentage of classes detected sorted by class size, again for $n=10$. (The figure shows class sizes up through size 24; this size limit includes 99.08% of all of the $n=10$ classes, even though the maximum class size is $n!$, for the complete graph.) The distribution shows an interesting pattern, with the most common class having a single member, with sizes 2, 3, and 4 being the next most common, but then dropping off quickly. As the inverse of the 0.267 ratio is simply the mean of this distribution, or 3.75, it is not surprising to see most of the distribution's weight appearing for the smallest class sizes. For $n=10$, the total mean class size including all class sizes is 3.731. Using just the first 24 classes shown in Figure 3, the mean is 3.344, showing that the classes in the long tail of the full distribution have only a small effect on the average class size. The pattern of the distribution shows that certain sizes appear more frequently than others. In particular, larger compound numbers occur more often than larger prime numbers. This is probably due to separate sets of undirected edges in the essential graph acting independently to produce class sizes that are products of the sizes of their independent components. The pattern shown in Figure 3 is not a fluke or simply noise: chi-squared differences between the 24-class distributions for $n=10$ and those for $n=9$, 8, 7, and 6 are all less than 0.005. Thus, the distribution appears to have already closely approached an asymptotic limit. Though it is not apparent from the figure, the count of class size 11 is identically zero, but none of the others shown are. However, many other class sizes greater than 24 have no counts, so 11 is simply the first one where this appears (for $n=10$). By considering the graph on $n$ vertices using $n$-1 edges that connects the vertices together in a single chain, it can be seen that, when no immoralities occur, this graph will produce a class of size $n$. Thus every class size will eventually have some classes present in this distribution if $n$ becomes large enough.

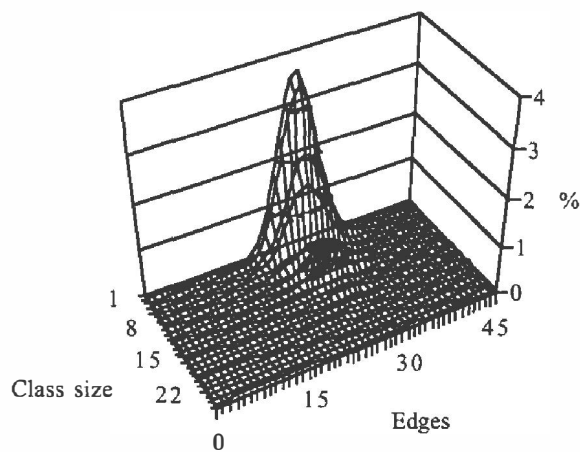

Figure 4: Classes by Sizes and Edges ($n$=10; sizes 1-24)

Finally, Figure 4 shows a mesh plot of equivalence class percentages distributed according to both class size and number of edges for $n$=10. The plot suggests that the two characteristics are acting independently; that is, that the distributions by edges are approximately Gaussian for any class size and the distribution of Figure 3 roughly occurs for any number of edges.

Table 2: Maxima Per Skeleton

| $n$ | v-Configurations | Classes |
|---|---|---|
| 1 | 0 | 1 |
| 2 | 0 | 1 |
| 3 | 1 | 2 |
| 4 | 4 | 6 |
| 5 | 9 | 22 |
| 6 | 18 | 104 |
| 7 | 30 | 594 |
| 8 | 48 | 3978 |
| 9 | 70 | 30768 |
| 10 | 100 | 257694 |

The program also kept track of the maximum number of v-configurations as well as the maximum number of classes produced by any single skeleton. Table 2 shows these values. It can be seen that the maximum number of v-configurations is indeed generally less than $n(n-1)(n-2)/6$. In fact, the maximum occurs on the evenly divided complete bipartite graphs and thus can easily be calculated. After algebraic simplification, the result is $(n-2)/2*\text{floor}(n/2)*\text{ceil}(n/2)$. The maximum number of classes occurs on a more complex (and as yet unpredicted) set of complete multipartite graphs. For $n \le 6$, both maxima occur on the same complete bipartite graphs, but for $n > 6$ the class maxima occur on graphs having more than two partitions of the vertices. While the maximum number of classes also remains unpredicted, it generally appears to be bounded above by $(n-1)!$.

## 4 IMPLICATIONS AND FUTURE DIRECTIONS

Clearly, with an average equivalence class size less than four, large computational advantages seem unlikely to be achieved by working directly with Markov equivalence classes. On the other hand, since the average class size is so small, it may be reasonable for algorithms to consider examining all of the members of a class during a model search. The distribution of classes by size may also indicate directions for future algorithms, since the vast majority of the class sizes are less than or equal to 24, and even smaller size limits may be of use. Finally, the Gaussian distribution of classes by edges may be of use in predicting future sampling efforts.

We hope that our results will stimulate further theoretical investigations into these and other questions concerning the combinatorial properties of equivalence classes of ADGs and other types of graphical Markov models.

### Acknowledgements

We thank David Madigan for access to the source code of his program to count equivalence classes for $n \le 5$.

This research supported in part by the United States National Science Foundation.